\documentclass{article}

\usepackage[export]{adjustbox}
\usepackage{amsmath}
\usepackage{arydshln}
\usepackage{authblk}
\usepackage{booktabs}
\usepackage[sort&compress]{cleveref}
\usepackage{fullpage}
\usepackage{amsthm,amsmath,amsfonts}
\usepackage{algorithmic}
\usepackage{algorithm}
\usepackage{bm}
\usepackage{bbm}
\usepackage{color,graphicx}
\usepackage[numbers,sort]{natbib}
\usepackage{enumerate}
\usepackage{multirow}
\usepackage{cases}
\usepackage{subfigure}
\usepackage{amssymb}

\newtheorem{theorem}{Theorem}
\theoremstyle{definition}
\numberwithin{equation}{section}

\title{A New Knowledge Gradient-based Method for Constrained Bayesian Optimization}

\author[1]{Wenjie Chen}
\author[1]{Shengcai Liu}
\author[1]{Ke Tang}

\affil[1]{Guangdong Provincial Key Laboratory of
	Brain-inspired Intelligent Computation, Department of Computer Science and
	Engineering, Southern University of Science and Technology, Shenzhen, China}

\date{}

\begin{document}

\maketitle

\begin{abstract}
	Black-box problems are common in real life like structural design, drug experiments, and machine learning. When optimizing black-box systems, decision-makers always consider multiple performances and give the final decision by comprehensive evaluations. Motivated by such practical needs, we focus on constrained black-box problems where the objective and constraints lack known special structure, and evaluations are expensive and even with noise. We develop a novel constrained Bayesian optimization approach  based on the knowledge gradient method ($c-\rm{KG}$). A new acquisition function is proposed to determine the next batch of samples considering optimality and feasibility. An unbiased estimator of the gradient of the new acquisition function is derived to implement the $c-\rm{KG}$ approach.
\end{abstract}

\section{Introduction}
\label{sect:intro}
Complex systems optimization is a critical challenge in real production and also the hot spot of academic research. The key factors that raise systems' complexity include (but are not limited to): inestimable structures, computationally intensive evaluations, stochastic noise, and multiple key performance indicators (KPIs). A typical example is a simulation-based optimization for an emergency department. Suppose we aim to optimize the patients' flow cost and departments' closeness by determining the corridors' widths via a simulation model. Due to the characteristics of the simulation model, there exists no explicit expression of the input and output, and the estimations are time-consuming and noise-corrupted. Furthermore, the multi-level performance indicators also lay a burden on optimization problems. In such a case, researchers always formulate a stochastically constrained black-box optimization problem where the primary performance is optimized subject to secondary performance constraints.

One of the methods to solve expensive black-box global optimization is  Bayesian optimization (BO). In recent decades, it has a great success in applications such as materials design \cite{frazier2016bayesian}, drug discovery \cite{negoescu2011knowledge}, reinforcement learning \cite{brochu2010tutorial}, automatic machine learning \cite{bergstra2011algorithms}, etc. BO also shows the versatile ability to solve different types of black-box optimization problems, such as noisy optimization problems \cite{letham2019constrained}, multi-objective optimization  problems \cite{belakaria2019max}, and multi-fidelity optimization problems \cite{song2019general}. Dating back to the seminal work of Schonlau $et\ al.$ \cite{schonlau1998global}, a general problem with constraints for a black-box system has gradually brought more attention from experts,
\begin{equation}\label{constrained BO}
	\begin{aligned}
		&\min\limits_{{\bf{x}}\in \Theta}\hspace{2ex} g_0({\bf{x}})\\
		&s.t.\hspace{3ex}
		g_i({\bf{x}})\leqslant 0 \hspace{4ex}\emph{i}=\rm{1},\rm{2},...,\emph{m}
	\end{aligned}
\end{equation}
where the domain $\Theta\subseteq\mathbb{R}^d$ with $d$ dimensions. The functions in the objective and constraints are derivative-free and expensive-to-evaluate. The evaluations of functions can be noise-corrupted or noise-free.

Compared with the fruitful work on unconstrained Bayesian optimization, the approaches to solving constrained problems are relatively few, especially the approaches to considering stochastic constraints \citep{gelbart2014bayesian,letham2019constrained}. In this paper, we concentrate on the stochastically constrained Bayesian optimization problems. Based on the knowledge gradient method, a new acquisition function is proposed to strike a balance between exploration and exploitation considering optimality and feasibility. Our contributions are twofold:
\begin{enumerate}
	\item We intend to deal with the stochastically constrained Bayesian optimization problem that has yet to be deeply explored. To guide an efficient sampling direction, a novel acquisition function called $c-\rm{KG}$ is developed. $c-\rm{KG}$ retains the original form of the knowledge gradient method and also contains the information of feasibility. It is an early attempt to exploit the potentiality of the knowledge gradient method for stochastically constrained BO problems.
	\item The next batch of samples is determined by the inner optimization problem that maximizes $c-\rm{KG}$. Here, we suggest a multi-start stochastic gradient ascent method and develop an unbiased estimator of the gradient of $c-\rm{KG}$ according to the mixture of infintiesimal perturbation analysis (IPA) and likelihood ratio (LR) methods.
\end{enumerate}

The remaining of the paper is organized as follows. Section \ref{section 2} introduces the background of Bayesian optimization. Section \ref{section 3} reviews the start-of-art work on constrained Bayesian optimization and points out the motivation of our method. In Section \ref{section 4}, the $c-\rm{KG}$ method is presented, followed by an efficient way to optimize $c-\rm{KG}$. Remarks and conclusions are discussed in Section \ref{section 5}.

\section{Background on Bayesian Optimization}\label{section 2}
BO is a sequential surrogate-model-based method to address global optimization problems. Fundamentally, it develops the surrogate models based on Bayesian statistics and determines the next move using a Bayesian interpretation of these surrogates. There are two critical ingredients in BO: a statistical function and an acquisition function. The former ingredient is used to build a prior distribution on the unknown function based on our belief and specify the mechanism of data generation. The commonly used model is Gaussian Process (GP). The later ingredient works as a loss function that decides where to sample the next point in the light of observations. 

We now yield a detailed discussion about these two components in Section \ref{Section 2.1} and Section \ref{Section 2.2}. Without loss of generality, we consider an unconstrained black-box problem as follows,
\begin{equation}\label{e3.1}
	\min\limits_{{\bf{x}}\in \Theta} f({\bf{x}})
\end{equation}
where $\Theta$ is usually a compact set of $\mathbb{R}^d$. Furthermore, we assume that the black-box function $f$ lacks the known structure and can be evaluated at any point ${\bf{x}}$.

\subsection{Gaussian Process}\label{Section 2.1}
GP regression is an efficient method to model the unknown smoothly varying function \cite{rasmussen2003gaussian}. With a finite collection of sample paths, the GP model is built and used to predict the mean and the variance analytically at any query point. GP is fully characterized by a particular mean function $\mu({\bf{x}})$ and a covariance function $K({\bf{x}},{\bf{x}}^{\prime})$. A sample $f({\bf{x}})$ at an arbitrary point ${\bf{x}}$ follows a Gaussian distribution: $f({\bf{x}})\sim \mathcal{GP}(\mu({\bf{x}}),K({\bf{x}},{\bf{x}}^{\prime}))$. 

To build the GP regression, we first place the prior on any sequence of points ${\bf{x}}^{(1:k)}:=\{{\bf{x}}^{(1)},...,{\bf{x}}^{(k)}\}$, resulting in the prior distribution on $f({\bf{x}}^{(1:k)}):=\{f({\bf{x}}^{(1)}),...,f({\bf{x}}^{(k)})\}$ as follows,
\begin{equation}
	f({\bf{x}}^{(1:k)})\sim\mathcal{N}(\mu({\bf{x}}^{(1:k)}),K({\bf{x}}^{(1:k)},{\bf{x}}^{(1:k)}))
\end{equation}
where the mean vector and the covariance matrix are
\begin{equation*}
	\begin{aligned}
		&\mu({\bf{x}}^{(1:k)}):=[\mu({\bf{x}}^{(1)}),...,\mu({\bf{x}}^{(k)})];\\
		&K({\bf{x}}^{(1:k)}):=[K({\bf{x}}^{(1)},{\bf{x}}^{(1)}),...,K({\bf{x}}^{(1)},{\bf{x}}^{(k)});,,,;K({\bf{x}}^{(k)},{\bf{x}}^{(1)}),...,K({\bf{x}}^{(k)},{\bf{x}}^{(k)})].
	\end{aligned}
\end{equation*}

Suppose we have a collection of initial observations $\mathcal{D}_n:=\{{\bf{x}}^{(i)},y^{(i)}\}_{i=1}^{n}$ where the observations can be noise-corrupted, that is, $y=f({\bf{x}})+\epsilon({\bf{x}})$ with $\epsilon({\bf{x}})\sim \mathcal{N}(0,{\sigma_{err}^2({\bf{x}})})$ where $\sigma^2_{err}({\bf{x}})$ represents the variance of measurement errors. In the GP regression, we have the prior on $f({\bf{x}}^{(1:n)})|{\bf{x}}^{(1:n)}\sim \mathcal{N}(\mu({\bf{x}}^{(1:n)}),K({\bf{x}}^{(1:n)},{\bf{x}}^{(1:n)}))$ and $y^{(1:n)}|f({\bf{x}}^{(1:n)}),{\sigma}_{err}^2({\bf{x}}^{(1:n)})\sim\mathcal{N}(f({\bf{x}}^{(1:n)}),{\sigma}_{err}^2({\bf{x}}^{(1:n)}))$ if we consider the random noise in the observation $y$. Given those observations, we may compute the posterior probability distribution of $f({\bf{x}})$ at a new point ${\bf{x}}^{(n+1)}$ following the Bayes' rule,
\begin{equation}
	y^{(n+1)}|{\bf{x}}^{(n+1)},\mathcal{D}_n\sim\mathcal{N}(\mu^{(n)}({\bf{x}}^{(n+1)}),(\sigma^{(n)}({\bf{x}}^{(n+1)}))^2)
\end{equation}
where
\begin{equation}\label{posterior dist}
	\begin{aligned}
		\mu^{(n)}({\bf{x}}^{(n+1)})=&\mu({\bf{x}}^{(n+1)})+K({\bf{x}}^{(n+1)},{\bf{x}}^{(1:n)})\{K({\bf{x}}^{(1:n)},{\bf{x}}^{(1:n)})\\
		&+{\rm{diag}}[\sigma_{err}^2({\bf{x}}^{(1)}),...\sigma_{err}^2({\bf{x}}^{(n)})]\}^{-1}[y^{(1:n)}-\mu({\bf{x}}^{(1:n)})]\\
		(\sigma^{(n)}({\bf{x}}^{(n+1)}))^2=&K({\bf{x}}^{(n+1)},{\bf{x}}^{(n+1)})-K({\bf{x}}^{(n+1)},{\bf{x}}^{(1:n)})\{K({\bf{x}}^{(1:n)},{\bf{x}}^{(1:n)})\\
		&+{\rm{diag}}[\sigma_{err}^2({\bf{x}}^{(1)}),...\sigma_{err}^2({\bf{x}}^{(n)})]\}^{-1}K({\bf{x}}^{(1:n)},{\bf{x}}^{(n+1)}).
	\end{aligned}
\end{equation}

We now elaborate on how to choose the mean function and kernel function.
\subsubsection*{Mean Function}
In the GP regression, the mean function provides an offset at any point, which implicates our prior belief in the unknown function. Often, the mean function is set to be a constant, that is, $\mu({\bf{x}})=\mu$. For convenience and generality, we assume the prior mean to be zero. 

Alternative priors can also be choices when the user has domain knowledge of the implicit function. For example, if we deem that $f({\bf{x}})$ has application-oriented structure or a trend, we can apply the mean function $\mu({\bf{x}})=\mu+\sum_{i=1}^{p}\beta_i\Phi_i({\bf{x}})$ where $\Phi_i({\bf{x}})$ is often a polynomial function of ${\bf{x}}$. 

\subsubsection*{Kernel Function}
In the GP regression, the kernel function intends to determine the smoothness and amplitude for the Gaussian process by grasping the connections between points. To achieve this goal, it should satisfy two fundamental properties. Firstly, the more close ${\bf{x}}_1$ and ${\bf{x}}_2$ are, the larger the positive correlation will be. This assumption is in line with our belief that closer points should have similar performance. Secondly, the covariance matrix should be positive semi-definite no matter how we choose the collection of points. 

A simple and popular choice for kernel is the $Gaussian$ kernel (aka the $squared\ exponential$ kernel),
\begin{equation*}
	K({\bf{x}},{\bf{x}}^\prime)=\alpha_0\exp(-||{\bf{x}}_1-{\bf{x}}_2||^2)
\end{equation*}
where  $||{\bf{x}}_1-{\bf{x}}_2||^2=\sum^d_{i=1}\alpha_i(x_{1,i}-x_{2,i})^2$, and $\alpha_{0:d}$ are the hyperparameters of the $Gaussian$ kernel. Users can vary the hyperparameters to create a diverse belief about the impact of ${\bf{x}}$ on $f({\bf{x}})$. Intuitively, the smaller value of $\alpha_i$ leads to a less violent fluctuation of $f({\bf{x}})$ in the $i$th dimension and less dependent on $i$th input, automatically discarding the irrelevant dimensions. 

Another common kernel is the $Mat\acute{e}rn$ kernel that is a flexible class of kernels parameterized by a smoothness parameter $\nu$,
\begin{equation*}
	K({\bf{x}},{\bf{x}}^\prime)=\frac{2^{1-\nu}}{\Gamma(\nu)}\left(\sqrt{2\nu}||{\bf{x}}-{\bf{x}}^\prime||\right)^\nu H_\nu\left(\sqrt{2\nu}||{\bf{x}}-{\bf{x}}^\prime||\right),
\end{equation*}
where $\Gamma(\cdot)$ and $H_{\nu}(\cdot)$ are the Gamma function and the modified Bessel function. Note that the smoothness parameter $\nu$ determines the differentiable times of a $Mat\acute{e}rn$ kernel. Namely, the samples from a GP can be differentiated $\nu-1$ times \cite{williams2006gaussian}. It is also interesting that when $\nu$ goes to infinity, the $Mat\acute{e}rn$ kernel is transformed to the version of the squared exponential kernel.

Choosing an appropriate kernel function requires the knowledge of engineering and automatic model selection. In addition to the commonly used $Gaussian$ kernel and $Mat\acute{e}rn$ kernels, there are other options examined in the machine learning literature \cite{williams2006gaussian}.

\subsection{Acquisition Function}\label{Section 2.2}
Thus far, we have presented the statistical model in GP. With the posterior model, we need an exploitation-exploration mechanism to guide the sequential search, in other words, how to select the next point ${\bf{x}}^{(n+1)}$ given the previous observations $\mathcal{D}_n$ and the prior belief about the unknown function $f({\bf{x}})$. We measure the benefit gained by sampling the next point with an acquisition function. By maximizing the acquisition function, we can make full use of the available information to quickly reach the optimal solution. The well-known acquisition functions mainly include expected improvement (EI), probability of improvement (PI), upper confidence bound (UCB), entropy search/predictive entropy search (ES/PES), and knowledge gradient (KG).

\subsubsection*{Probability of Improvement}
The probability of improvement intends to measure the probability of improvement over the incumbent $f({\bf{x}}^*)$ where ${\bf{x}}^*=\arg\min_{{\bf{x}}\in{\bf{x}}^{(1:n)}} f({\bf{x}})$ as follows,
\begin{equation}\label{original PI}
	\begin{aligned}
		\alpha_{PI}({\bf{x}},\mathcal{D}_n)&:=P(f({\bf{x}})\leqslant f({\bf{x}}^*))={\rm{\Phi}}\left(\frac{f({\bf{x}}^*)-\mu^{(n)}({\bf{x}})}{\sigma^{(n)}({\bf{x}})}\right)
	\end{aligned}
\end{equation}
where ${\rm{\Phi}}({\bf{x}})$ is the normal cumulative distribution function.  

It is noticed that the formulation in \eqref{original PI} is pure exploitation. 
An alternative PI is proposed by adding a tunable parameter $\epsilon$, written as
\begin{equation*}
	\begin{aligned}
		\alpha_{PI}({\bf{x}},\mathcal{D}_n)&:=P(f({\bf{x}})\leqslant f({\bf{x}}^*)-\epsilon)={\rm{\Phi}}\left(\frac{f({\bf{x}}^*)-\mu^{(n)}({\bf{x}})-\epsilon}{\sigma^{(n)}({\bf{x}})}\right).
	\end{aligned}
\end{equation*}

We can control the desired improvement over the current best observation and select the point most likely to provide at least $\epsilon$ improvement by tuning the parameter $\epsilon$.

\subsubsection*{Upper/Lower Confidence Bound}
The upper confidence bound criterion \cite{srinivas2009gaussian} is one of the popular methods to balance the exploitation and exploration in Bayesian optimization, often with provable cumulative regret bounds. The guiding principle behind this strategy is to combine the expected performance and uncertainty into the acquisition function with the following formulation,
\begin{equation*}
	\alpha_{UCB}({\bf{x}},\mathcal{D}_n):=\mu^{(n)}({\bf{x}})+\beta\sigma^{(n)}({\bf{x}}),
\end{equation*}
where the tunable parameter $\beta$ works as $\epsilon$ in PI.

\subsubsection*{Expected Improvement}
Dating back to the pioneering work of Jones $et\ al.$ \cite{jones1998efficient}, expected improvement is one of the most commonly used acquisition functions. The underlying idea of EI is how to make a final decision to return an optimal solution from all the sampled points with noise-free evaluations. EI measures the gain of the final decision by incorporating both the probability and the magnitude of the improvement into the expectation as the following expression,
\begin{equation*}
	\alpha_{EI}({\bf{x}},\mathcal{D}_n):={\rm{E}}[max\{0,f({\bf{x}}^*)-f({\bf{x}})\}|\mathcal{D}_n]
\end{equation*}
where ${\bf{x}}^*=\arg\min_{{\bf{x}}\in{\bf{x}}^{(1:n)}} f({\bf{x}})$. 

EI can also be evaluated in closed-form in terms of the normal cumulative distribution function ${\rm{\Phi}}({\bf{x}})$ and the normal probability density function $\varphi({\bf{x}})$, so that
\begin{equation*}
	\begin{aligned}
		\alpha_{EI}({\bf{x}},\mathcal{D}_n)&:={\rm{E}}[max\{0,f({\bf{x}}^*)-f({\bf{x}})\}|\mathcal{D}_n]\\
		&=(f({\bf{x}}^*)-\mu^{(n)}({\bf{x}})){\rm{\Phi}}\left(\frac{f({\bf{x}}^*)-\mu^{(n)}({\bf{x}})}{\sigma^{(n)}({\bf{x}})}\right)+\sigma^{(n)}({\bf{x}})\varphi\left(\frac{f({\bf{x}}^*)-\mu^{(n)}({\bf{x}})}{\sigma^{(n)}({\bf{x}})}\right).
	\end{aligned}
\end{equation*}

\subsubsection*{Entropy Search and Predictive Entropy Search}
Entropy search \cite{hennig2012entropy} 
records the information on the location of the best solution according to its differential entropy that implies the uncertainty, seeking to maximize information about the location ${\bf{x}}^*$. The entropy reduction  can be written as,
\begin{equation*}
	\alpha_{ES}({\bf{x}},\mathcal{D}_n):={\rm{H}}(P_n({\bf{x}}^*|\mathcal{D}_n))-{\rm{E}}_{f({\bf{x}})}[{\rm{H}}(P_n({\bf{x}}^*|\mathcal{D}_n,{\bf{x}},f({\bf{x}})))]
\end{equation*}
where ${\rm{H}}(P_n({\bf{x}}^*)|\mathcal{D}_n)$ denotes the entropy of the time-$n$ posterior distribution on ${\bf{x}}^*$ given the observations $\mathcal{D}_n$, and ${\rm{H}}(P_n({\bf{x}}^*|\mathcal{D}_n,{\bf{x}},f({\bf{x}})$ represents the entropy of the time-$n+1$ posterior distribution on ${\bf{x}}^*$ given the observations $\mathcal{D}_n$ and  $f({\bf{x}})$ at ${\bf{x}}$.

Predictive Entropy Search \cite{hernandez2014predictive} provides another way to measure the differential entropy as follows,
\begin{equation*}
	\alpha_{PES}({\bf{x}},\mathcal{D}_n):={\rm{H}}(P_n(f({\bf{x}})|\mathcal{D}_n))-{\rm{E}}_{{\bf{x}}^*}[{\rm{H}}(P_n(f({\bf{x}})|\mathcal{D}_n,{\bf{x}}^*))]
\end{equation*}
where the reduction in the entropy of ${\bf{x}}^*$ caused by the observation $f({\bf{x}})$ is equal to the decrease in the entropy of $f({\bf{x}})$ caused by observing ${\bf{x}}^*$ because of the symmetry of mutual information between $f({\bf{x}})$ and ${\bf{x}}^*$. 
The approximations of ES and PES require different computational techniques, resulting in different sampling decisions.

\subsubsection*{Knowledge Gradient}
Knowledge gradient differs from the expected improvement method in the criterion to determine the current best solution. KG requires the decision-maker's tolerance to risk so that a final recommendation with uncertainty is accepted. In other words, the final solution can be any point, even if it has not been sampled before. If we go for risk-neutrality \cite{berger2013statistical}, the final point can be returned in terms of the expected value $\mu({\bf{x}})$, i.e., ${\bf{x}}^*=\arg\min_{{\bf{x}}\in\Theta}\mu^{(n)}({\bf{x}})$ where $\mu^{(n)}({\bf{x}})$ is the conditional expected value of $f({\bf{x}})$. The subtle difference results in the ability of the KG method to handle noisy evaluations. 

Frazier $et\ al.$ \citep{frazier2008knowledge,frazier2009knowledge} introduced the knowledge gradient method for Gaussian process regression over a discrete solution space. Later works \citep{scott2011correlated,wu2016parallel,wu2017discretization} generalized KG to solve continuous optimization problems. We now write the expression of KG as follows,
\begin{equation*}
	\alpha_{KG}({\bf{x}},\mathcal{D}_n):={\rm{E}}[\mu^{(n)}_*-\mu^{(n+1)}_*|\mathcal{D}_n,{\bf{x}}^{(n+1)}={\bf{x}}]
\end{equation*}
where $\mu^{(n+1)}_*=\min_{{\bf{x}}\in\Theta}\mu^{(n+1)}({\bf{x}})$, and $\mu^{(n)}_*$ is defined in the similar way. 

We summarize the general framework in Algorithm \ref{BO}.

\begin{algorithm}
	\caption{Generic pseudo-code for Bayesian Optimization}   
	\label{BO}
	\begin{algorithmic}[1]  
		\REQUIRE The number of initial samples $n_0$, the number of total iterations $N$, the index of iterations $n$. 
		\ENSURE The estimated best feasible solution $\hat{{\bf{x}}}^*$.
		\STATE Place a GP prior (the mean function $\mu({\bf{x}})$ and the kernel function $K({\bf{x}},{\bf{x}}^{\prime})$) on $f({\bf{x}})$.
		\STATE According to a design of experiment method, sample $n_0$ initial points ${\bf{x}}^{(1:n_0)}:=\{{\bf{x}}^{(1)},...,{\bf{x}}^{(n_0)}\}$, and obtain their observations $y^{(1:n_0)}$. Set $n=n_0$.
		\STATE Update the GP posterior probability distribution based on the observations.
		\WHILE {$n\leqslant N$}
		\STATE Find ${\bf{x}}^{(n+1)}$ by optimizing the acquisition function over the GP using the current posterior distribution. 
		\STATE Obtain the observation of ${\bf{x}}^{(n+1)}$, retrain the GP model, and get the estimated best feasible solution $\hat{{\bf{x}}}^*$. 
		\STATE $n=n+1$.
		\ENDWHILE
	\end{algorithmic}  
\end{algorithm} 

\section{Related Work on Constrained Bayesian Optimization}\label{section 3}

Most constrained Bayesian optimization methods are based on the variants of the expected improvement acquisition function. Schonlau $et\ al.$ \cite{schonlau1998global} and Gardner $et\ al.$ \cite{gardner2014bayesian} paid attention to the problem in \eqref{constrained BO} with noiseless evaluations and assumed independence between $g_0({\bf{x}})$ and $g_i({\bf{x}})$ for any query point. The guiding principle is using a generalized improvement subject to constraints,
\begin{equation*}
	\begin{aligned}
		I_{\mathcal{D}_n}(\mathbf{x})&=\left\{\begin{array}{ll}
			g_0(\mathbf{x}^*)-g_0(\mathbf{x})& \text { if } g_0(\mathbf{x})<g_0(\mathbf{x}^*) \text { and } g_i({\bf{x}})\leqslant 0 \text { for } i=1, \ldots,m \\
			0 & \text { otherwise }
		\end{array}\right.\\
		&=\Delta({\bf{x}})\max\left\{0,g_0(\mathbf{x}^*)-g_0(\mathbf{x})\right\}
	\end{aligned}
\end{equation*}
where ${\bf{x}}^*=\arg\min_{{\bf{x}}\in{{\bf{x}}^{(1:n)}}} g_0({\bf{x}})\text { s.t. } g_i({\bf{x}})\leqslant 0\ (i=1, \ldots,m )$. The quantity $\Delta({\bf{x}})$ is a Bernoulli random variable with the parameter about the probability $\rm{Pr}\{{\bf{x}}\text{ is a feasible solution}\}$.

We then derive the generalized expected improvement with the assumption that $g_0({\bf{x}})$ and $g_i({\bf{x}})$ are independent,

\begin{equation}\label{EIC}
	\begin{aligned}
		\alpha_{EIC}({\bf{x}},\mathcal{D}_n)&:={\rm{E}}\left[I_{\mathcal{D}_n}(\mathbf{x})|\mathcal{D}_n,{\bf{x}}^{(n+1)}={\bf{x}}\right]\\
		&={\rm{E}}\left[\Delta({\bf{x}})\max\left\{0,g_0(\mathbf{x}^*)-g_0(\mathbf{x})\right\}|\mathcal{D}_n,{\bf{x}}^{(n+1)}={\bf{x}}\right]\\
		&=\alpha_{EI}({\bf{x}},\mathcal{D}_n)\prod\limits_{i=1}^m{\rm{Pr}}[g_i({\bf{x}})\leqslant 0]
	\end{aligned}
\end{equation}
where $\prod_{i=1}^m{\rm{Pr}}[g_i({\bf{x}})\leqslant 0]$ can be regarded as the weight on the improvement of the objective value.

When the noise presents in evaluations, the previous methods mentioned above will fail. Gelbart $et\ al.$ \cite{gelbart2014bayesian} considered the case that there existed noise in the constraints and assumed that the objective and constraints were evaluated independently. The best feasible solution is selected according to its objective value with the constraint that each constraint should be satisfied with a probability over $1-\delta_i$ where $\delta_i$ is a user-specified value. The resulting constraint-weighted expected improvement is similar to \eqref{EIC}. When there is no feasible solution in some iteration, the acquisition function is designed to find the feasible region according to the probability $\prod_{i=1}^m{\rm{Pr}}[g_i({\bf{x}})]$. Letham $et\ al.$ \cite{letham2019constrained} tackled the constrained Bayesian optimization problem with the stochastic objective and stochastic constraints and proposed the noisy expected improvement criterion under greedy batch optimization. Inspired by the methods in \cite{schonlau1998global} and \cite{gardner2014bayesian}, they defined a new utility function and placed a penalty for not having a feasible solution. Therefore, the objective value is also used to determine the improvement when there is no feasible solution in the set of sampled points, which is different from the method in \cite{gelbart2014bayesian}. The resulting improvement function is written as,
\begin{equation}\label{new IF}
	I_{\mathcal{D}_n}(\mathbf{x})=\left\{\begin{array}{ll}
		g_0(\mathbf{x}^*)-g_0(\mathbf{x})& \text { if } g_0(\mathbf{x})<g_0(\mathbf{x}^*) \text { and } g_i({\bf{x}})\leqslant 0 \text { for } i=1, \ldots,m  \text { and } S_n\neq\emptyset\\
		M-g_0(\mathbf{x})& \text { if } g_i({\bf{x}})\leqslant 0 \text { for } i=1, \ldots,m \text { and } S_n=\emptyset\\
		0 & \text { otherwise }\\
	\end{array}\right.
\end{equation}
where $S_n$ is the set of feasible observations. Let $g^{(1:n)}_{i}({\bf{x}}):=\{g_i({\bf{x}}^{(1)}),...,g_i({\bf{x}}^{(n)})\}\ (i=0, \ldots,m)$ denote the performance values at the observations. Because $g^{(1:n)}_{i}({\bf{x}})$ is unknown in the noise setting, the expectation in  \eqref{new IF}  is extended to noisy cases by integrating in terms of $g^{(1:n)}_{i}({\bf{x}})$.

Gramacy $et\ al.$ \cite{gramacy2016modeling} applied the augmented Lagrangian (AL) to integrate the constraints into the objective function. They deployed the GP surrogate model to each component in the augmented Lagrangian, resulting in a nonanalytic expected improvement function that can be estimated by Monte Carlo integration over the posterior. Picheny $et\ al.$ \cite{picheny2016bayesian} introduced an alternative slack variable augmented Lagrangian. Under the slack-AL-based formulation, the new expected improvement function can be easily calculated after reparameterization via quadrature.

Another popular way to handle the constrained Bayesian optimization is to revisit the entropy search/predictive entropy search. Often, EI-based methods encounter the failure modes to determine the current best feasible solution if all the sampled solutions are infeasible or the evaluations are corrupted by noise. However, information-based approaches can handle such issues because they record information about the best solution's location. Hernández-Lobato $et\ al.$ \cite{hernandez2015predictive} extended PES to noisy constrained problems and assumed the independent GP priors of the objective and constraint functions. They proposed Predictive Entropy Search with Constraints (PESC) with the purpose of approximating the expected information gain about the location of the constrained global optimum ${\bf{x}}^*$ whose posterior distribution is $p({\bf{x}}^*|\mathcal{D}_n)$. Based on \cite{hernandez2015predictive}, Hernández-Lobato $et\ al.$ \cite{hernandez2016general} further solved problems with decoupled constraints, in which subsets of the objective and constraint functions may be evaluated independently. Perrone $et\ al.$ \cite{perrone2019constrained} concentrated on Max-value Entropy Search (MES) \cite{wang2017max}, a simple and efficient alternative to PES, and proposed constrained Max-value Entropy Search (cMES) that could handle both continuous and binary constraints.

In addition to the above constrained Bayesian optimization approaches, other methods provide us new ways to solve constrained BO problems \citep{bernardo2011optimization,ariafar2019admmbo,ghoreishi2019multi,tran2019pbo,pourmohamad2019statistical,picheny2014stepwise}. Knowledge gradient has inherent advantages in dealing with noisy evaluations, and has been extended to batch optimization \cite{wu2016parallel} but not to constrained BO yet because it is nontrivial to integrate the constraints into the KG while retaining the tractability of the expectation. To the best of our knowledge, the proposed constrained knowledge gradient ($c-{\rm{KG}}$) method is the early attempt to apply KG to solve constrained BO.

\section{Constrained Knowledge Gradient}\label{section 4}

We propose a novel acquisition function called constrained knowledge gradient ($c-\rm{KG}$) for Bayesian global optimization.  We place the Gaussian process prior on the performance $g_i({\bf{x}})$ specified by the mean function $\mu_i({\bf{x}})$ and the kernel function $K_i({\bf{x}}_1,{\bf{x}}_2)$ for $i=0,1,...,m$. 

Suppose we have sampled $n$ points ${\bf{x}}^{(1:n)}:=\{{\bf{x}}^{(1)},{\bf{x}}^{(2)},...,{\bf{x}}^{(n)}\}$ and obtained their observations $G_i({\bf{x}}^{(1:n)})$ for $i=0,1,...,m$ where the evaluation $G_i({\bf{x}})|g_i({\bf{x}})\sim N(g_i({\bf{x}}),\sigma_{err,i}^2({\bf{x}}))$ and $\sigma_{err,i}^2({\bf{x}})$ represents the variance of measurement errors. Then, the posterior distribution is still the Gaussian process with the mean function $\mu_i^{(n)}({\bf{x}})$ and the kernel function $K_i^{(n)}({\bf{x}}_1,{\bf{x}}_2)$ which satisfy 
\begin{equation*}
	\begin{aligned}
		\mu_i^{(n)}({\bf{x}})=&\mu_i({\bf{x}})+K_i({\bf{x}},{\bf{x}}^{(1:n)})(K_i({\bf{x}}^{(1:n)},{\bf{x}}^{(1:n)})\\
		&+{\rm{diag}}\{\sigma_{err,i}^2({\bf{x}}^{(1)}),...,\sigma_{err,i}^2({\bf{x}}^{(n)})\})^{-1}(G_i({\bf{x}}^{(1:n)})-\mu_i({\bf{x}}^{(1:n)}))\\
		K_i^{(n)}({\bf{x}}_1,{\bf{x}}_2)=&K_i({\bf{x}}_1,{\bf{x}}_2)-K_i({\bf{x}}_1,{\bf{x}}^{(1:n)})(K_i({\bf{x}}^{(1:n)},{\bf{x}}^{(1:n)})\\
		&+{\rm{diag}}\{\sigma_{err,i}^2({\bf{x}}^{(1)}),...,\sigma_{err,i}^2({\bf{x}}^{(n)})\})^{-1}K_i({\bf{x}}^{(1:n)},{\bf{x}}_2).
	\end{aligned}
\end{equation*}

In the knowledge gradient method, the decision-maker is risk-neutral and selects the best solution with the optimal expected value, even though it may not be observed. To extend the knowledge gradient method for the stochastically constrained problem, it is natural to choose the best feasible solution  $\widehat{{\bf{x}}}_*^{(n)}=\arg\min_{{\bf{x}}\in \Theta}\mu^{(n)}_0({\bf{x}})\ {\rm{s.t.}}\mu^{(n)}_i({\bf{x}})\leqslant 0$ for $i=1,2,...,m$. 

If we were to sample one more batch with $q$ points  ${\bf{z}}^{(1:q)}:=\{{\bf{z}}^{(1)},{\bf{z}}^{(2)},...,{\bf{z}}^{(q)}\}$ and obtained their measurements $G^{(n)}_i({\bf{z}}^{(1:q)})$ for $i=0,1,...,m$, we have  $G^{(n)}_i({\bf{z}}^{(j)})\sim N(\mu^{(n)}_i({\bf{z}}^{(j)}),(\sigma^{(n)}_i({\bf{z}}^{(j)}))^2)$ for $j=1,...,q$ where 
\begin{equation*}
	\begin{aligned}
		\mu_i^{(n)}({\bf{z}}^{(j)})=&\mu_i({\bf{z}}^{(j)})+K_i({\bf{z}}^{(j)},{\bf{x}}^{(1:n)})(K_i({\bf{x}}^{(1:n)},{\bf{x}}^{(1:n)})\\
		&+{\rm{diag}}\{\sigma_{err,i}^2({\bf{x}}^{(1)}),...,\sigma_{err,i}^2({\bf{x}}^{(n)})\})^{-1}(G_i({\bf{x}}^{(1:n)})-\mu_i({\bf{x}}^{(1:n)}))\\
		(\sigma^{(n)}_i({\bf{z}}^{(j)}))^2=&K_i({\bf{z}}^{(j)},{\bf{z}}^{(j)})-K_i({\bf{z}}^{(j)},{\bf{x}}^{(1:n)})(K_i({\bf{x}}^{(1:n)},{\bf{x}}^{(1:n)})\\
		&+{\rm{diag}}\{\sigma_{err,i}^2({\bf{x}}^{(1)}),...,\sigma_{err,i}^2({\bf{x}}^{(n)})\})^{-1}K_i({\bf{x}}^{(1:n)},{\bf{z}}^{(j)}).
	\end{aligned}
\end{equation*}

Then, we would get new posterior distributions for $g_i({\bf{x}})$ with the mean function $\mu_i^{(n+q)}({\bf{x}})$ and the covariance function $K_i^{(n+q)}({\bf{x}}_1,{\bf{x}}_2)$ and update the best feasible solution $\widehat{{\bf{x}}}_*^{(n+q)}=\arg\min_{{\bf{x}}\in \Theta}\mu^{(n+q)}_0({\bf{x}})\ {\rm{s.t.}}\mu^{(n+q)}_i({\bf{x}})\leqslant 0$ for $i=1,2,...,m$. 
\begin{equation}\label{eq:2}
	\begin{aligned}
		\mu_i^{(n+q)}({\bf{x}})=&\mu^{(n)}_i({\bf{x}})+K^{(n)}_i({\bf{x}},{\bf{z}}^{(1:q)})(K^{(n)}_i({\bf{z}}^{(1:q)},{\bf{z}}^{(1:q)})\\
		&+{\rm{diag}}\{\sigma_i^2({\bf{z}}^{(1)}),...,\sigma_i^2({\bf{z}}^{(q)})\})^{-1}(G^{(n)}_i({\bf{z}}^{(1:q)})-\mu^{(n)}_i({\bf{z}}^{(1:q)}))\\
		K_i^{(n+q)}({\bf{x}}_1,{\bf{x}}_2)=&K^{(n)}_i({\bf{x}}_1,{\bf{x}}_2)-K^{(n)}_i({\bf{x}}_1,{\bf{z}}^{(1:q)})(K^{(n)}_i({\bf{z}}^{(1:q)},{\bf{z}}^{(1:q)})\\
		&+{\rm{diag}}\{\sigma_i^2({\bf{z}}^{(1)}),...,\sigma_i^2({\bf{z}}^{(q)})\})^{-1}K^{(n)}_i({\bf{z}}^{(1:q)},{\bf{x}}_2).
	\end{aligned}
\end{equation}

However, it is non-trivial to integrate the constraints into the KG's framework. In our method, we aim to retain the basic form of KG and add information about the feasibility of the $q$ points. Assuming the measurements among $m$ constraints are independent, we have the  following acquisition function $c-{\rm{KG}}({\bf{z}}^{(1:q)})$,
\begin{equation}
	\begin{aligned}
		c-{\rm{KG}}({\bf{z}}^{(1:q)})&=\exp\left\{{\rm{E}}\left[\min\limits_{{\bf{x}}\in \Theta^{(n)}_f}\mu^{(n)}_0({\bf{x}})-\min\limits_{{\bf{x}}\in \Theta^{(n+q)}_f}\mu^{(n+q)}_0({\bf{x}})|{\bf{z}}^{(1:q)}\right]\right\}\cdot \prod\limits_{j=1}^q\prod\limits_{i=1}^m{\rm{Pr}}\left[G^{(n)}_i({\bf{z}}^{(j)})\leqslant 0\right]\\
		&=\exp\left\{{\rm{E}}\left[\min\limits_{{\bf{x}}\in \Theta^{(n)}_f}\mu^{(n)}_0({\bf{x}})-\min\limits_{{\bf{x}}\in \Theta^{(n+q)}_f}\mu^{(n+q)}_0({\bf{x}})|{\bf{z}}^{(1:q)}\right]\right\}\cdot \prod\limits_{j=1}^q\prod\limits_{i=1}^m{\rm{E}}\left[1\left\{G^{(n)}_i({\bf{z}}^{(j)})\leqslant 0\right\}\right]
	\end{aligned}
\end{equation}
where $\Theta^{(n)}_f=\{{\bf{x}}\in\Theta: \mu^{(n)}_i({\bf{x}})\leqslant 0,\ \forall i\in\{1,...,m\}\}$. $\Theta^{(n+q)}_f$ is defined in the same way.

The $c-{\rm{KG}}({\bf{z}}^{(1:q)})$ integrates the optimality and feasibility in the acquisition function, retaining expectation's tractability. The first part measures the improvement on the best feasible objective values where the exponential function guarantees the positive value of the improvement. The second part takes the feasibility of ${\bf{z}}^{(1:q)}$. Under the framework of Bayesian optimization, the closer designs tends to perform more similarly. Therefore, the best feasible solution is more likely to be close to other feasible solutions. In this way, the more improvement on the best feasible objective value and the more likely of being feasible result in the larger value of $c-{\rm{KG}}({\bf{z}}^{(1:q)})$.

For the constrained Bayesian optimization, we maximize the acquisition function to determine the next batch ${\bf{z}}^{(1:q)}$ as follows,
\begin{equation}\label{c-KG}
	\max\limits_{{\bf{z}}^{(1:q)}\in \Theta}\hspace{2ex} c-{\rm{KG}}({\bf{z}}^{(1:q)}).
\end{equation}

\subsection{The Computation of $c-{\rm{KG}}$}
Following \eqref{eq:2}, 
$\mu_i^{(n+q)}({\bf{x}})=\mu^{(n)}_i({\bf{x}})+K^{(n)}_i({\bf{x}},{\bf{z}}^{(1:q)})(K^{(n)}_i({\bf{z}}^{(1:q)},{\bf{z}}^{(1:q)})+{\rm{diag}}\{\sigma_i^2({\bf{z}}^{(1)}),...,$\\$\sigma_i^2({\bf{z}}^{(q)})\})^{-1}(G^{(n)}_i({\bf{z}}^{(1:q)})-\mu^{(n)}_i({\bf{z}}^{(1:q)}))$. Because $G^{(n)}_i({\bf{z}}^{(1:q)})-\mu^{(n)}_i({\bf{z}}^{(1:q)})$ is normally distributed with zero mean and covariance matrix $K^{(n)}_i({\bf{z}}^{(1:q)},{\bf{z}}^{(1:q)})+{\rm{diag}}\{\sigma_i^2({\bf{z}}^{(1)}),...,\sigma_i^2({\bf{z}}^{(q)})\}$ with respect to the
posterior.
$\mu^{(n+q)}_i({\bf{x}})$  can be rewritten as \cite{wu2017discretization}
\begin{equation}\label{eq:4}
	\mu^{(n+q)}_i({\bf{x}})=\mu^{(n)}_i({\bf{x}})+{\tilde{\sigma}}_i^{(n)}({\bf{x}},{\bf{z}}^{(1:q)}){\textit{\textbf{Z}}}_{q,i}.
\end{equation}

In equation \eqref{eq:4}, ${\textit{\textbf{Z}}}_{q,i}$ is a standard $q$-dimensional normal random vector for the $i$th performance, and ${\tilde{\sigma}}_i^{(n)}({\bf{x}},{\bf{z}}^{(1:q)})=K_i^{(n)}({\bf{x}},{\bf{z}}^{(1:q)})(D_i^{(n)}({\bf{z}}^{(1:q)})^T)^{-1}$ where $D_i^{(n)}({\bf{z}}^{(1:q)})$ is the Cholesky factor of the covariance matrix $K^{(n)}_i({\bf{z}}^{(1:q)},{\bf{z}}^{(1:q)})+{\rm{diag}}\{\sigma_{i}^2({\bf{z}}^{(1)}),...,\sigma_{i}^2({\bf{z}}^{(q)})\}$. To compute the $c-{\rm{KG}}$ factor for the batch ${\bf{z}}^{(1:q)}$, we can sample ${\textit{\textbf{Z}}}_{q,i}$, solve the optimization problems $\min_{{\bf{x}}\in \Theta^{(n)}_f}\mu^{(n)}_0({\bf{x}})$ and $\min_{{\bf{x}}\in \Theta^{(n+q)}_f}\mu^{(n+q)}_0({\bf{x}})$ by constrained nonlinear programming, then plug in \eqref{eq:4}, repeat many times, and take the average.

\subsection{The Maximization of $c-{\rm{KG}}$}
We use the stochastic approximation to maximize $c-{\rm{KG}}({\bf{z}}^{(1:q)})$ and provide an unbiased estimator of the following gradient of the $c-{\rm{KG}}$ factor,
\begin{equation}\label{eq:5}
	\frac{\partial }{\partial z_{t,k}}c-{\rm{KG}}({\bf{z}}^{(1:q)})=\frac{\partial}{\partial z_{t,k}}\left\{\exp\left\{{\rm{E}}\left[L({\bf{z}}^{(1:q)},{\textit{\textbf{Z}}}_{q,0})\right]\right\}\cdot \prod\limits_{j=1}^q\prod\limits_{i=1}^m {\rm{E}}\left[1\left\{G^{(n)}_i({\bf{z}}^{(j)})\leqslant 0\right\}\right]\right\}
\end{equation}
where $z_{t,k}$ is the $k$th dimension of the $t$th point in ${\bf{z}}^{(1:q)}$, and $L({\bf{z}}^{(1:q)},{\textit{\textbf{Z}}}_{q,0})=\min_{{\bf{x}}\in \Theta^{(n)}_f}\mu^{(n)}_0({\bf{x}})-$\\$\min_{{\bf{x}}\in \Theta^{(n+q)}_f}\left(\mu^{(n)}_0({\bf{x}})+{\tilde{\sigma}}_0^{(n)}({\bf{x}},{\bf{z}}^{(1:q)}){\textit{\textbf{Z}}}_{q,0}\right)$.

We require two assumptions for later analysis. 
\begin{enumerate}
	\item The mean function $\mu(\cdot)$ and the kernel function $K(\cdot)$ are continuously differentiable. 
	\item The domain $\Theta$ is compact.
\end{enumerate}

Many mean functions and kernel functions satisfy the assumption 1 and assumption 2, for example the constant mean function and the Gaussian kernel function. 

Because of the assumption 1, $\mu^{(n+q)}_i({\bf{x}})$ remains continuous differentiability after the multiplication, the inverse, and the Cholesky operator given ${\textit{\textbf{Z}}}_{q,i}$. Then, $L({\bf{z}}^{(1:q)},{\textit{\textbf{Z}}}_{q,0})$ and its expectation are continuously differentiable.
Because $G^{(n)}_i({\bf{z}}^{(j)})\sim N(\mu_i^{(n)}({\bf{z}}^{(j)}),(\sigma_i^{(n)}({\bf{z}}^{(j)}))^2)$, ${\rm{E}}\left[1\left\{G^{(n)}_i({\bf{z}}^{(j)})\leqslant 0\right\}\right]={\rm{Pr}}\left[G^{(n)}_i({\bf{z}}^{(j)})\leqslant 0\right]=\Phi\left(-\mu_i^{(n)}({\bf{z}}^{(j)})/\sigma_i^{(n)}({\bf{z}}^{(j)})\right)$ which also has continuous differentiability. According to the product rule, we express the gradient of $c-{\rm{KG}}({\bf{z}}^{(1:q)})$ as follows,
\begin{equation}\label{eq:6}
	\begin{aligned}
		\frac{\partial }{\partial z_{t,k}}c-{\rm{KG}}({\bf{z}}^{(1:q)})&=\frac{\partial}{\partial z_{t,k}}\left\{\exp\left\{{\rm{E}}\left[L({\bf{z}}^{(1:q)},{\textit{\textbf{Z}}}_{q,0})\right]\right\}\cdot \prod\limits_{j=1}^q\prod\limits_{i=1}^m {\rm{E}}\left[1\left\{G^{(n)}_i({\bf{z}}^{(j)})\leqslant 0\right\}\right]\right\}\\
		&=\frac{\partial}{\partial z_{t,k}}{\rm{E}}\left[L({\bf{z}}^{(1:q)},{\textit{\textbf{Z}}}_{q,0})\right]\cdot\exp\left\{{\rm{E}}\left[L({\bf{z}}^{(1:q)},{\textit{\textbf{Z}}}_{q,0})\right]\right\}\cdot \prod\limits_{j=1}^q\prod\limits_{i=1}^m {\rm{E}}\left[1\left\{G^{(n)}_i({\bf{z}}^{(j)})\leqslant 0\right\}\right]\\
		&\quad+\sum\limits_{i_0=1}^m\frac{\partial}{\partial z_{t,k}}{\rm{E}}\left[1\left\{G^{(n)}_{i_0}({\bf{z}}^{(t)})\leqslant 0\right\}\right]\cdot\exp\left\{{\rm{E}}\left[L({\bf{z}}^{(1:q)},{\textit{\textbf{Z}}}_{q,0})\right]\right\}\\
		&\quad\quad\cdot\left\{{\rm{Pr}}\left[G^{(n)}_{i_0}({\bf{z}}^{(t)})\leqslant 0\right]\right\}^{-1}\cdot\prod\limits_{j=1}^q\prod\limits_{i=1}^m {\rm{Pr}}\left[G^{(n)}_i({\bf{z}}^{(j)})\leqslant 0\right].
	\end{aligned}
\end{equation}

Under the regularity assumptions above, we will prove that the interchange of expectation and differentiation is allowed in \eqref{eq:6} and derive an unbiased estimator for the gradient of $c-{\rm{KG}}({\bf{z}}^{(1:q)})$ by applying a mixture of infinitesimal perturbation analysis (IPA) and likelihood ratio (LR) methods.

$G^{(n)}_{i_0}({\bf{z}}^{(t)})$ can be written as $G^{(n)}_{i_0}({\bf{z}}^{(t)},W_{i_0}^{(t)})=\mu_{i_0}^{(n)}({\bf{z}}^{(t)})+\sigma_{i_0}^{(n)}({\bf{z}}^{(t)})W_{i_0}^{(t)}$ where $\mu_{i_0}^{(n)}({\bf{z}}^{(t)})$ and $\sigma_{i_0}^{(n)}({\bf{z}}^{(t)})$ depend on ${\bf{z}}^{(t)}$ explicitly and $W_{i_0}^{(t)}$ is a standard normal random variable. We define a new random variable $Y_{i_0}^{(t)}=G^{(n)}_{i_0}({\bf{z}}^{(t)},W_{i_0}^{(t)})$ whose support is $\mathbb{R}$ and probability density function is $f_{Y_{i_0}^{(t)}}(y)=n(d)/\sigma_{i_0}^{(n)}({\bf{z}}^{(t)})$ where $n(\cdot)$ is the probability density of a standard normal random variable and $d=(y-\mu_{i_0}^{(n)}({\bf{z}}^{(t)}))/\sigma_{i_0}^{(n)}({\bf{z}}^{(t)})$. For ease of notation, we write $W_{i_0}^{(t)}$ and $Y_{i_0}^{(t)}$ as $W$ and $Y$, respectively. Let $\mathbb{P}_{Y}$ be the probability measure induced by $Y$ and $\hat{\mathbb{Q}}$ be the Lebesgue measure. Therefore, we have
\begin{equation}\label{eq:7}
	\begin{aligned}
		\frac{\partial}{\partial z_{t,k}}{\rm{E}}\left[1\left\{G^{(n)}_{i_0}({\bf{z}}^{(t)},W)\leqslant 0\right\}\right]&=\frac{\partial}{\partial z_{t,k}}\int_{\Omega_W}1\left\{G^{(n)}_{i_0}({\bf{z}}^{(t)},W)\leqslant 0\right\}d\mathbb{P}_W\\
		&=\frac{\partial}{\partial z_{t,k}}\int_{\Omega_Y}1\left\{Y\leqslant 0\right\}d\mathbb{P}_Y({\bf{z}}^{(t)})\\
		&=\frac{\partial}{\partial z_{t,k}}{\rm{E}}_{\mathbb{P}_{Y}}\left[1\left\{Y\leqslant 0\right\}\right]\\
		&=\frac{\partial}{\partial z_{t,k}}{\rm{E}}_{\hat{\mathbb{Q}}}\left[1\left\{Y\leqslant 0\right\}\frac{d\mathbb{P}_{Y}}{d\hat{\mathbb{Q}}}(Y)\right].
	\end{aligned}
\end{equation}
where $W$ is defined on a probability space $(\Omega_W,\mathcal{F}_W,\mathbb{P}_W)$, and $Y$ is defined on a probability space $(\Omega_Y,\mathcal{F}_Y,\mathbb{P}_Y)$. Here, $\Omega_W$ and $\Omega_Y$ are sample spaces, and $\mathcal{F}_W$ and $\mathcal{F}_Y$ are $\sigma$--algebra defined on $\Omega_W$ and $\Omega_Y$, respectively. We write $\mathbb{P}_Y({\bf{z}}^{(t)})$ to indicate that $\mathbb{P}_Y$ depends on ${\bf{z}}^{(t)}$.

Combining \eqref{eq:6} and \eqref{eq:7}, it is noticed that there are two types of terms that depend on $z_{t,k}$ explicitly: $L({\bf{z}}^{(1:q)},{\textit{\textbf{Z}}}_{q,0})$ and the Radon-Nikodym derivative $({d\mathbb{P}_{Y}}/{d\hat{\mathbb{Q}}})(Y)$ of $\mathbb{P}_{Y}$ with respect to $\mathbb{Q}$ ($({d\mathbb{P}_{Y}}/{d\hat{\mathbb{Q}}})(Y)$ corresponds to the probability density function $f_{Y}(y)$ in our problem). The indicator function $1\left\{Y\leqslant 0\right\}$ no longer depends on the parameter $z_{t,k}$. Therefore, given $Y$, a perturbation in $z_{t,k}$ will have an impact on the function $L({\bf{z}}^{(1:q)},{\textit{\textbf{Z}}}_{q,0})$ and $({d\mathbb{P}_{Y}}/{d\hat{\mathbb{Q}}})(Y)$. We now apply IPA to $L({\bf{z}}^{(1:q)},{\textit{\textbf{Z}}}_{q,0})$. Assuming differentiation and expectation can be interchanged, we have
\begin{equation}
	\begin{aligned}
		\frac{\partial}{\partial z_{t,k}}{\rm{E}}\left[L({\bf{z}}^{(1:q)},{\textit{\textbf{Z}}}_{q,0})\right]&={\rm{E}}\left[\frac{\partial}{\partial z_{t,k}}L({\bf{z}}^{(1:q)},{\textit{\textbf{Z}}}_{q,0})\right]
	\end{aligned}
\end{equation}
where
\begin{equation}
	\begin{aligned}
		{\mathcal{D}}_0&\triangleq\frac{\partial }{\partial z_{t,k}}L({\bf{z}}^{(1:q)},{\textit{\textbf{Z}}}_{q,0})\\
		&=-\frac{\partial }{\partial z_{t,k}}\min_{{\bf{x}}\in \Theta^{(n+q)}_f}\left(\mu^{(n)}_0({\bf{x}})+{\tilde{\sigma}}_0^{(n)}({\bf{x}},{\bf{z}}^{(1:q)}){\textit{\textbf{Z}}}_{q,0}\right)\\
		&=-\frac{\partial }{\partial z_{t,k}}{\tilde{\sigma}}_0^{(n)}({\bf{x}}^*({\textit{\textbf{Z}}}_{q,0}),{\bf{z}}^{(1:q)}){\textit{\textbf{Z}}}_{q,0}\\
		&=-\left\{\left(\frac{\partial }{\partial z_{t,k}}K_0^{(n)}({\bf{x}}^*({\textit{\textbf{Z}}}_{q,0}),{\bf{z}}^{(1:q)})\right)(D_0^{(n)}(z^{(1:q)})^T)^{-1}\right.\\
		&\left.\quad\quad\quad-K_0^{(n)}({\bf{x}}^*({\textit{\textbf{Z}}}_{q,0}),{\bf{z}}^{(1:q)})(D_0^{(n)}(z^{(1:q)})^T)^{-1}\left(\frac{\partial }{\partial z_{t,k}}D_0^{(n)}(z^{(1:q)})^T\right)(D_0^{(n)}(z^{(1:q)})^T)^{-1}\right\}\cdot {\textit{\textbf{Z}}}_{q,0}
	\end{aligned}
\end{equation}
where ${\bf{x}}^*({\textit{\textbf{Z}}}_{q,0})=\arg\min_{{\bf{x}}\in \Theta^{(n+q)}_f} \mu^{(n)}_0({\bf{x}})+{\tilde{\sigma}}_0^{(n)}({\bf{x}},{\bf{z}}^{(1:q)}){\textit{\textbf{Z}}}_{q,0}$.

To apply LR to $({d\mathbb{P}_{Y}}/{d\hat{\mathbb{Q}}})(Y)$, we define $\mathbb{Q}$ to be the probability measure such that $\mathbb{P}_{Y}$ is absolutely continuous with respect to $\mathbb{Q}$. In particular, we set $\mathbb{Q}$ to be the probability measure induced by $Y$. With the assumption that differentiation and expectation can be interchanged, we have 
\begin{equation}
	\begin{aligned}
		\frac{\partial}{\partial z_{t,k}}{\rm{E}}_{\hat{\mathbb{Q}}}\left[1\left\{Y\leqslant 0\right\}\frac{d\mathbb{P}_{Y}}{d\hat{\mathbb{Q}}}(Y)\right]&=\frac{\partial}{\partial z_{t,k}}{\rm{E}}_{{\mathbb{Q}}}\left[1\left\{Y\leqslant 0\right\}\frac{d\mathbb{P}_{Y}}{d{\mathbb{Q}}}(Y)\right]\\
		&={\rm{E}}_{{\mathbb{Q}}}\left[1\left\{Y\leqslant 0\right\}\frac{\partial\ln(d\mathbb{P}_{Y}/d\hat{\mathbb{Q}})(Y)}{\partial z_{t,k}}\frac{d\mathbb{P}_{Y}}{d\mathbb{Q}}(Y)\right]
	\end{aligned}
\end{equation}
where
\begin{equation}\label{Dtj0}
	\begin{aligned}
		{\mathcal{D}}_{(t,i_0)}&\triangleq 1\left\{Y\leqslant 0\right\}\cdot\frac{\partial\ln(d\mathbb{P}_{Y}/d\hat{\mathbb{Q}})(Y)}{\partial z_{t,k}}\frac{d\mathbb{P}_{Y}}{d\mathbb{Q}}(Y)\\
		&=1\left\{Y\leqslant 0\right\}\cdot\frac{\partial\ln(d\mathbb{P}_{Y}/d\hat{\mathbb{Q}})(Y)}{\partial z_{t,k}}\\
		&=1\left\{G^{(n)}_{i_0}({\bf{z}}^{(t)})\leqslant 0\right\}\cdot\left\{\frac{\left[G^{(n)}_{i_0}({\bf{z}}^{(t)})-\mu_{i_0}^{(n)}({\bf{z}}^{(t)})\right]}{\left[\sigma_{i_0}^{(n)}({\bf{z}}^{(t)})\right]^2}\cdot\frac{\partial \mu_{i_0}^{(n)}({\bf{z}}^{(t)})}{{\partial z_{t,k}}}\right.\\
		&\left.\quad+\frac{\left[G^{(n)}_{i_0}({\bf{z}}^{(t)})-\mu_{i_0}^{(n)}({\bf{z}}^{(t)})\right]^2-\left[\sigma_{i_0}^{(n)}({\bf{z}}^{(t)})\right]^2}{\left[\sigma_{i_0}^{(n)}({\bf{z}}^{(t)})\right]^3}\cdot\frac{\partial \sigma_{i_0}^{(n)}({\bf{z}}^{(t)})}{{\partial z_{t,k}}}\right\}.\\
	\end{aligned}
\end{equation}

Through a reverse change of variables, the last equality in \eqref{Dtj0} is obtained and written in terms of the original random variable $G^{(n)}_{i_0}({\bf{z}}^{(t)})$.

\begin{theorem}
	Under the assumptions 1--2, the unbiased estimator of $\partial c-{\rm{KG}}({\bf{z}}^{(1:q)})/\partial z_{t,k}$ is given by 
	\begin{equation*}\label{eq:10}
		\begin{aligned}
			{\mathcal{D}}=&{\mathcal{D}}_0\cdot\exp\left\{{\rm{E}}\left[L({\bf{z}}^{(1:q)},{\textit{\textbf{Z}}}_{q,0})\right]\right\}\cdot\prod\limits_{j=1}^q\prod\limits_{i=1}^m {\rm{Pr}}\left[G^{(n)}_i({\bf{z}}^{(j)})\leqslant 0\right]\\
			&+\sum\limits_{i_0=1}^m{\mathcal{D}}_{(t,i_0)}\cdot\exp\left\{{\rm{E}}\left[L({\bf{z}}^{(1:q)},{\textit{\textbf{Z}}}_{q,0})\right]\right\}\cdot\left\{{\rm{Pr}}\left[G^{(n)}_{i_0}({\bf{z}}^{(t)})\leqslant 0\right]\right\}^{-1}\cdot\prod\limits_{j=1}^q\prod\limits_{i=1}^m {\rm{Pr}}\left[G^{(n)}_i({\bf{z}}^{(j)})\leqslant 0\right].
		\end{aligned}
	\end{equation*}
\end{theorem}

\textbf{\emph{Proof}}.
Under the assumption 1,  for any fixed ${\textit{\textbf{Z}}}_{q,0}$, $L({\bf{z}}^{(1:q)},{\textit{\textbf{Z}}}_{q,0})$ are absolutely continuous and differentiable almost everywhere. 
\begin{equation*}\label{eq:10}
	\begin{aligned}
		\left|\frac{\partial }{\partial z_{t,k}}L({\bf{z}}^{(1:q)},{\textit{\textbf{Z}}}_{q,0})\right|&=\left|-\frac{\partial }{\partial z_{t,k}}{\tilde{\sigma}}_0^{(n)}({\bf{x}}^*({\textit{\textbf{Z}}}_{q,0}),{\bf{z}}^{(1:q)}){\textit{\textbf{Z}}}_{q,0}\right|\\
		&=\left|-\left\{\left(\frac{\partial }{\partial z_{t,k}}K_0^{(n)}({\bf{x}}^*({\textit{\textbf{Z}}}_{q,0}),{\bf{z}}^{(1:q)})\right)(D_0^{(n)}(z^{(1:q)})^T)^{-1}\right.\right.\\
		&\left.\left.\quad-K_0^{(n)}({\bf{x}}^*({\textit{\textbf{Z}}}_{q,0}),{\bf{z}}^{(1:q)})(D_0^{(n)}(z^{(1:q)})^T)^{-1}\left(\frac{\partial }{\partial z_{t,k}}D_0^{(n)}(z^{(1:q)})^T\right)(D_0^{(n)}(z^{(1:q)})^T)^{-1}\right\}\cdot{\textit{\textbf{Z}}}_{q,0}\right|.
	\end{aligned}
\end{equation*}

Let $U_0({\bf{z}}^{(1:q)})=K^{(n)}_0({\bf{z}}^{(1:q)},{\bf{z}}^{(1:q)})+{\rm{diag}}\{\sigma_{0}^2({\bf{z}}^{(1)}),...,\sigma_{0}^2({\bf{z}}^{(q)})\}$. Under the assumption 1, $U_0({\bf{z}}^{(1:q)})$ and $\partial U_0({\bf{z}}^{(1:q)})/\partial z_{t,k}$ are continuous. As the Cholesky factor of $U_0({\bf{z}}^{(1:q)})$, $(D_0^{(n)}(z^{(1:q)})^T)^{-1}$ and $\partial D_0^{(n)}(z^{(1:q)})^T/\partial z_{t,k}$ also remain continuity. Therefore, ${\partial }{\tilde{\sigma}}_0^{(n)}({\bf{x}}^*({\textit{\textbf{Z}}}_{q,0}),{\bf{z}}^{(1:q)})/\partial z_{t,k}$ is continuous. As $\Theta$ is compact, ${\partial }{\tilde{\sigma}}_0^{(n)}({\bf{x}}^*({\textit{\textbf{Z}}}_{q,0}),{\bf{z}}^{(1:q)})/\partial z_{t,k}$ is bounded by a vector $0\leqslant \Gamma<\infty$. We have that $\left|{\partial}L({\bf{z}}^{(1:q)},{\textit{\textbf{Z}}}_{q,0})/\partial z_{t,k}\right|\leqslant\sum_{j=1}^{q}\Gamma_j|Z_{j,0}|$ where ${\textit{\textbf{Z}}}_{q,0}=(Z_{1,0},...,Z_{q,0})^T$ and ${\rm{E}}(\sum_{j=1}^q\Gamma_j|Z_{j,0}|)=\sqrt{2/\pi}\sum_{j=1}^q\Gamma_j<\infty$. 
According to the Theorem 1 in \cite{l1990unified}, We can get an unbiased estimator for ${\partial{\rm{E}}\left[L({\bf{z}}^{(1:q)},{\textit{\textbf{Z}}}_{q,0})\right]/\partial z_{t,k}}$.

Let $H({\bf{z}}^{(t)},Y)=1\left\{Y\leqslant 0\right\}(d\mathbb{P}_{Y}/d{\mathbb{Q}})(Y)$. We now prove $H({\bf{z}}^{(t)},Y)$ to be differentiable and continuous everywhere in $\Theta$ given any $Y$. View $Y$ as the set of values taken by a finite sequence of independent random variables, and set $\mathbb{Q}=\mathbb{P}_{Y}({\bf{z}}^{(t)}_0)$ where ${\bf{z}}^{(t)}_0\in\Theta$. In this case, 
\begin{equation*}
	\begin{aligned}
		H({\bf{z}}^{(t)},Y)&=1\left\{Y\leqslant 0\right\}\frac{d\mathbb{P}_{Y}}{d{\mathbb{Q}}}(Y)=1\left\{Y\leqslant 0\right\}\frac{(d\mathbb{P}_{Y}/d\hat{\mathbb{Q}})(Y)}{(d{\mathbb{Q}}/d\hat{\mathbb{Q}})(Y)}=1\left\{Y\leqslant 0\right\}\frac{f_{Y,{\bf{z}}^{(t)}}(y)}{f_{Y,{\bf{z}}^{(t)}_0}(y)}
	\end{aligned}
\end{equation*}
where $f_{Y}(y)=n(d)/\sigma_{i_0}^{(n)}({\bf{z}}^{(t)})$, $n(\cdot)$ is the probability density of a standard normal random variable, and $d=(y-\mu_{i_0}^{(n)}({\bf{z}}^{(t)}))/\sigma_{i_0}^{(n)}({\bf{z}}^{(t)})$. With assumption 1, for any fixed $Y$, $H({\bf{z}}^{(t)},Y)$ exists and is continuously differentiable in $\Theta$.

Let $G^{(n)}_{i_0}({\bf{z}}^{(t)})=\mu_{i_0}^{(n)}({\bf{z}}^{(t)})+\sigma_{i_0}^{(n)}({\bf{z}}^{(t)})W_{i_0}$, and add it in \eqref{Dtj0}. We have that
\begin{equation*}
	\begin{aligned}
		\left|\frac{\partial H({\bf{z}}^{(t)},Y)}{\partial z_{t,k}}\right|&=
		\left|1\left\{\mu_{i_0}^{(n)}({\bf{z}}^{(t)})+\sigma_{i_0}^{(n)}({\bf{z}}^{(t)})W_{i_0}\leqslant 0\right\}\frac{\partial\ln(f_{Y})}{\partial z_{t,k}}\right|\\
		&=\left|1\left\{\mu_{i_0}^{(n)}({\bf{z}}^{(t)})+\sigma_{i_0}^{(n)}({\bf{z}}^{(t)})W_{i_0}\leqslant 0\right\}\left(A({\bf{z}}^{(t)})W_{i_0}^2+B({\bf{z}}^{(t)})W_{i_0}+C({\bf{z}}^{(t)})\right)\right|
	\end{aligned}
\end{equation*}
where 
\begin{equation*}
	A({\bf{z}}^{(t)})=\frac{{\partial \sigma_{i_0}^{(n)}({\bf{z}}^{(t)})}/{{\partial z_{t,k}}}}{\sigma_{i_0}^{(n)}({\bf{z}}^{(t)})}, \quad B({\bf{z}}^{(t)})=\frac{{\partial \mu_{i_0}^{(n)}({\bf{z}}^{(t)})}/{{\partial z_{t,k}}}}{\sigma_{i_0}^{(n)}({\bf{z}}^{(t)})}, \quad {\rm{and}}\quad C({\bf{z}}^{(t)})=-\frac{{\partial \sigma_{i_0}^{(n)}({\bf{z}}^{(t)})}/{{\partial z_{t,k}}}}{\sigma_{i_0}^{(n)}({\bf{z}}^{(t)})}.
\end{equation*}

Under the assumption 1, $A({\bf{z}}^{(t)})$, $B({\bf{z}}^{(t)})$, and $C({\bf{z}}^{(t)})$ are continuous. Because $\Theta$ is a compact set, $A({\bf{z}}^{(t)})$, $B({\bf{z}}^{(t)})$, and $C({\bf{z}}^{(t)})$ are bounded by $\Gamma_A$, $\Gamma_B$, and $\Gamma_C$, respectively, where $0\leqslant\Gamma_A, \Gamma_B, \Gamma_C<\infty$. Similarly, we have that $\left|{\partial}H({\bf{z}}^{(t)},{\textit{\textbf{Z}}}_q)/\partial z_{t,k}\right|\leqslant\Gamma_AW_{i_0}^2+\Gamma_B|W_{i_0}|+\Gamma_C$ and ${\rm{E}}(\Gamma_AW_{i_0}^2+\Gamma_B|W_{i_0}|+\Gamma_C)=\Gamma_A+\sqrt{2/\pi}\Gamma_B+\Gamma_C<\infty$. Therefore, the Assumption 1 in \cite{l1990unified} is satisfied, and we get the unbiased estimator of ${\partial}{\rm{E}}\left[1\left\{G^{(n)}_{i_0}({\bf{z}}^{(t)})\leqslant 0\right\}\right]/\partial z_{t,k}$.  

We summarize the constrained BO method based on the $c-\rm{KG}$ acquisition function in Algorithm \ref{Chapter 3 Algorithm 2}.
\begin{algorithm}
	\caption{Framework of Bayesian optimization with $c-\rm{KG}$}   
	\label{Chapter 3 Algorithm 2}
	\begin{algorithmic}[1]  
		\REQUIRE The number of initial samples $n_0$, the number of a batch of samples $q$, the number of total iterations $N$, the index of iterations $n$. 
		\ENSURE The estimated best feasible solution $\hat{{\bf{x}}}^*$.
		\STATE Place a GP prior (the mean function $\mu({\bf{x}})$ and the kernel function $K({\bf{x}},{\bf{x}}^{\prime})$) on $f({\bf{x}})$ for the objective and constraints.
		\STATE According to a design of experiment method, sample $n_0$ initial points ${\bf{x}}^{(1:n_0)}:=\{{\bf{x}}^{(1)},{\bf{x}}^{(2)},...,{\bf{x}}^{(n_0)}\}$, and obtain their objective and constraint observations. 
		\STATE Update the GP posterior for each measurement using all the observations.
		\STATE If $\Theta^{(n)}_f\neq\emptyset$, solve $\min_{{\bf{x}}\in \Theta^{(n)}_f}\mu^{(n)}_0({\bf{x}})$ by the constrained nonlinear programming, and get the estimated best feasible solution $\hat{{\bf{x}}}^*$; otherwise, go to Step 2.	
		\WHILE {$n\leqslant N$}
		\STATE Solve problem \eqref{c-KG} to determine a batch of solutions ${\bf{x}}^{(1:q)}:=\{{\bf{x}}^{(1)},{\bf{x}}^{(2)},...,{\bf{x}}^{(q)}\}$ and get their observations.
		\STATE Retrain the Gaussian process models and get the estimated best feasible solution $\hat{{\bf{x}}}^*$. 
		\STATE $n=n+1$.
		\ENDWHILE
	\end{algorithmic}  
\end{algorithm} 

\section{Conclusions}\label{section 5}
In this study, we investigate the constrained Bayesian optimization problems and examine the potential of the knowledge gradient method to deal with the constrained problem. A novel acquisition function called constrained knowledge gradient ($c-\rm{KG}$) is developed to generalize KG for constrained problems with noisy evaluations. $c-\rm{KG}$ naturally integrates the information of feasibility to the acquisition function, retaining the expectation's tractability. To maximize $c-\rm{KG}$, we use the stochastic approximation method with an unbiased estimator of the gradient of $c-\rm{KG}$ based on IPA and LR. 

\bibliographystyle{unsrt}
\bibliography{Library}

\end{document}